\documentclass[conference]{IEEEtran}

\usepackage{times}
\usepackage{graphicx} % more modern
\graphicspath{{images/}}
\usepackage{algorithm}
\usepackage{algorithmic}
\usepackage{url}
\usepackage{subcaption}
\usepackage{wrapfig}
\usepackage{here}
\usepackage{cite}
\usepackage{amsmath}
\usepackage{array}
\usepackage{multirow}
\usepackage{boxedminipage}

\begin{document}
%
% paper title
% Titles are generally capitalized except for words such as a, an, and, as,
% at, but, by, for, in, nor, of, on, or, the, to and up, which are usually
% not capitalized unless they are the first or last word of the title.
% Linebreaks \\ can be used within to get better formatting as desired.
% Do not put math or special symbols in the title.
\title{Structured Sparse Ternary Weight Coding of\\ Deep Neural Networks for Efficient Hardware Implementations}

% author names and affiliations
% use a multiple column layout for up to three different
% affiliations
\author{\IEEEauthorblockN{Yoonho Boo and Wonyong Sung}
\IEEEauthorblockA{Department of Electrical Engineering and Computer Science\\
Seoul National University\\
Seoul 151-744, South Korea\\
Email: yhboo.research@gmail.com; wysung@snu.ac.kr}
}

% make the title area
\maketitle

% As a general rule, do not put math, special symbols or citations
% in the abstract
\begin{abstract}
Deep neural networks (DNNs) usually demand a large amount of operations for real-time inference. Especially, fully-connected layers contain a large number of weights, thus they usually need many off-chip memory accesses for inference. We propose a weight compression method for deep neural networks, which allows values of +1 or -1 only at predetermined positions of the weights so that decoding using a table can be conducted easily. For example, the structured sparse (8,2) coding allows at most two non-zero values among eight weights. This method not only enables multiplication-free DNN implementations but also compresses the weight storage by up to x32 compared to floating-point networks. Weight distribution normalization and gradual pruning techniques are applied to mitigate the performance degradation. The experiments are conducted with fully-connected deep neural networks and convolutional neural networks.

% Deep neural networks demand a huge amount of weights storage, which makes on-chip memory implementation difficult. Fixed-point quantization is one solution that reduces the cost of weights storage and computation. However, in the case of the large-sized networks, quantization alone cannot compress the weights storage at the on-chip memory level. We propose structured sparse fixed-point networks to reduce the size of the weights storage. The weights matrices are compressed into a structure look-up table and indexing addresses. In addition, Weights distribution normalization and gradual pruning technique are applied to prevent the performance degradation. The experiments are conducted with fully connected deep neural networks and convolutional neural networks.
\end{abstract}

% no keywords

% For peer review papers, you can put extra information on the cover
% page as needed:
% \ifCLASSOPTIONpeerreview
% \begin{center} \bfseries EDICS Category: 3-BBND \end{center}
% \fi
%
% For peerreview papers, this IEEEtran command inserts a page break and
% creates the second title. It will be ignored for other modes.
\IEEEpeerreviewmaketitle

\section{Introduction}
\label{sec:Intro}

Deep neural networks (DNNs) show high performance in various classification problems. However, the implementation of them requires a much increased number of arithmetic operations when compared to traditional pattern recognition and classification algorithms~\cite{bishop1995neural, hinton2012deep}.
 In particular, fully-connected layers of fully-connected deep neural networks (FCDNNs) or convolutional neural networks (CNN) contain a large number of parameters, which makes it difficult to implement them using resource-limited hardware. The number of weights usually exceeds millions, which incurs many off-chip memory accesses and large power consumption. When all the weights are expressed as ternary values (+1, 0, and -1), it is possible not only to reduce the off-chip memory access but also to remove multiplications.

Weight quantization is a straightforward way of reducing the size of parameters~\cite{hwang2014fixed, kim2014x1000, park2016fpga, li2016ternary, zhu2017trained, rastegari2016xnor, han2016deep}. Many DNNs show high resiliency on weight quantization\cite{sung2015resiliency}. In particular, the precision of the weight can be lowered to 2-bit (+1, 0 and -1) without much performance degradation by retraining the quantized networks~\cite{hwang2014fixed, li2016ternary, zhu2017trained}. This ternary representation can compress the weight by x16 when compared to the 32-bit floating-point network. In recent years, there were some trials to represent the DNN weight using the 1-bit binary (+1 and -1) format to increase the compression ratio (x32) and perform inference using only logical operations~\cite{rastegari2016xnor}. However, it still shows considerable performance degradation. Compression of DNNs and CNNs employing a few data compression techniques has been developed in ~\cite{han2016deep}. This method prunes small valued weights to remove 89\% and 92.5\% of connections for AlexNet~\cite{krizhevsky2012imagenet} and VGG-16~\cite{simonyan2015vgg} and applies vector quantization. Also the final weights are compressed using a compressed sparse row/compressed sparse column (CSC/CSR) format with relative index and Huffman coding to achieve x35, x49 times weight storage compression. However, not only the decoding method used in this approach is quite complex, but also the implementation of decompressed networks requires high-precision arithmetic units~\cite{han2016eie}.

In this work, we develop a decoding-conscious weight representation method not only to highly compress the network but also to implement it very efficiently in real-time. The proposed algorithm trains the network so that the weights are represented using a structured sparse ternary format. This format allows +1 or -1 only at specified locations, while most of the values are pruned to zero. The network can achieve the compression ratio of almost up to x32, but the performance of the network is much better than the binary or XNOR networks. The hardware for inference contains a small look-up table for decompressing the code, but the procedure is very simple and deterministic. The data-path needs a reduced number of arithmetic units because most of the weights are pruned to zero. The indexing addresses can be easily interpreted to corresponding weights. Also, this method has a good scalability because the look-up table size is independent of the network complexity. However, training the structured sparsity network is more difficult than optimizing the conventional ternary valued networks. We use batch normalization and weight normalization techniques to mitigate the performance degradation. Also gradual pruning technique is applied for a large-sized network to improve the performance. The proposed scheme was evaluated on FCDNN, VGG-9, and AlexNet and obtained the compression rate between x23 and x32.

%structure
\begin{figure*}[ht]
\centering
\includegraphics[height = 0.22\linewidth , width = 0.9\linewidth]{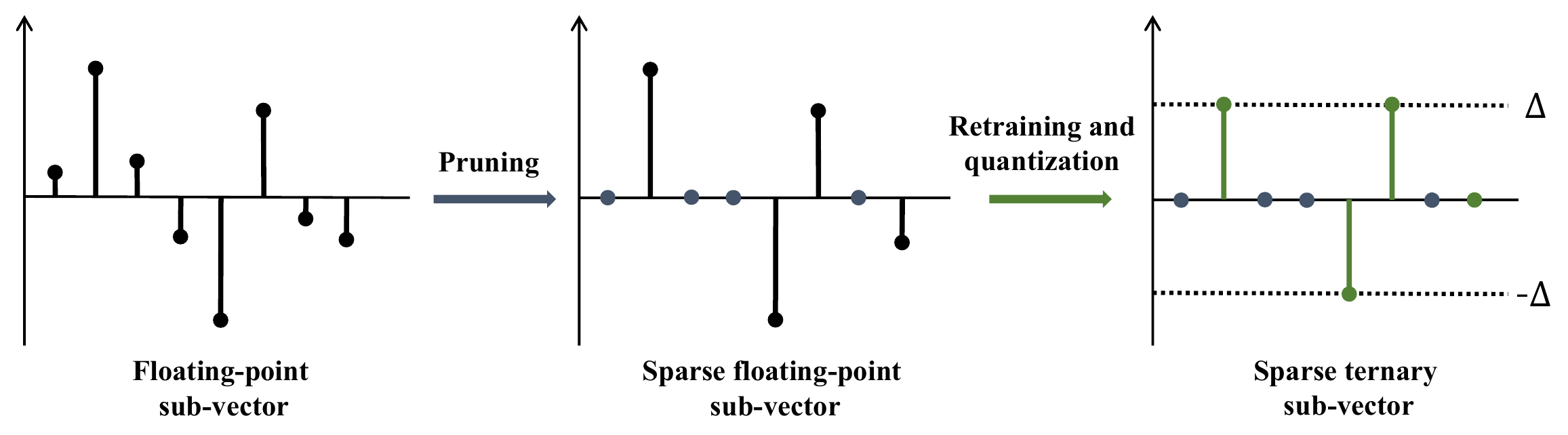}
\caption{The process of sub-vector structure sparse ternary quantization when $(N,K)$ is $(8,4)$.}
\label{fig:structure}
\end{figure*} 

The rest of this paper is organized as follows. Section~\ref{sec:SSFN} presents the proposed structured sparse ternary networks. In Section~\ref{sec:retraining}, the network training method is presented. The effects of DNN normalizers and gradual pruning are also explained. Experimental results are shown in Section~\ref{sec:experiments}. Concluding remarks follow in Section~\ref{sec:conclusion}.

\section{Structured sparse ternary quantization}
\label{sec:SSFN}
%In this section, we explain the structure and the weights storing method.

\subsection{Review of ternary quantization}
\label{subsec:review}
Ternary quantization represents a weight of a DNN using only +1, 0, and -1, and can achieve good performance by retraining~\cite{hwang2014fixed, li2016ternary, zhu2017trained}. In this case, one ternary weight is represented by 2 bits, thus a compression ratio of x16 can be obtained when compared with the 32-bit floating-point format. However, since only 3 levels (+1, 0, -1) are used, the information that can be represented with 2-bit per weight is not fully utilized.  In addition, the ternary optimization results show that very high portion, about 85\%, of the weights are zero, which implies the possibility of additional compression~\cite{hwang2014fixed}.

%compression
\begin{table}[]
\centering
\caption{The number of table entries ($T$), the table size($S_{T}$) and the length address ($I$) for one sub-vector.}
\label{compression}
{\renewcommand{\arraystretch}{2}

\setlength{\tabcolsep}{15pt}
\begin{tabular}{cccc}
\hline\hline
Codes  & $T$     & $S_{T}$ (KB) & $I$ (bits) \\ \hline
$(16,4)$ & 34113 & 136.452   & 16       \\ \hline
$(16,3)$ & 4993  & 19.972    & 13       \\ \hline
$(16,2)$ & 513   & 2.052     & 10       \\ \hline
$(8,2)$  & 129   & 0.258     & 8        \\ \hline
$(8,1)$  & 17    & 0.034     & 5        \\ \hline
$(4,1)$  & 9     & 0.009     & 4        \\ \hline\hline
\end{tabular}}
\end{table}

\subsection{Sub-vector structured sparsity}
\label{subsec:sub_vector}
The proposed structured ternary quantization divides a weight into many one-dimensional sub-vectors with the size of $N$, and each sub-vector is represented by a ternary vector with a limited number, $K$, of +1 or -1. We first prune each sub-vector of the floating-point weight matrix to have only $K$ non-zero values. Then, quantization and retraining are performed with pruned weights kept to zero. By the retraining and quantization, the number of non-zero values can be decreased. The process is described in~\figurename~\ref{fig:structure} when $(N,K)$ is $(8,4)$. In this figure, the final vector is determined as $[0, +1, 0, 0, -1, +1, 0, 0]$, and contains only 3 non-zero values

 This structure can reduce the weight storage using a look-up table and indexing addresses. For example, the $(4,1)$ structured sparse coding denotes that the sub-vector length is 4 and only one position is allowed to be +1 or -1. Since the number of sub-vectors satisfying this condition is 9 including $(0, 0, 0, 0)$, $(0, 0, 0, +1)$, $(0, 0, 0, -1)$, $...$, and $(-1, 0, 0, 0)$, the sub-vector index can be represented in 4 bits. As a result, the number of bits for (4, 1) structured ternary encoding is just 1-bit per weight excluding the memory for look-up table. In this structured sparse ternary encoding scheme, the sub-vector size, $N$, needs to be limited, such as 8 or 16, because the look-up table size increases as $N$ grows. The proposed structured sparse ternary network needs a look-up table and indexing addresses. All possible sub-vectors are stored in the look-up table. Since the number of look-up table entries, $T$, is $\sum_{i=0}^{K}\binom Ni 2^i$, the table occupies $2NT$ bits and the indexing address of the sub-vector demands $\lceil\log_{2}{T}\rceil$ bits. For example, if $(N,K)$ is $(16,4)$, the total number of table entries is 34,113 and the address length is 16 bits. Also, if $(N,K)$ is $(8,1)$, the address length becomes 5 bits and the table size is just 34 Bytes. Since the table size is relatively small, the weight storage can be further reduced when compared to the conventional ternary coding. The proposed method only performs table indexing without any complicated decoding process, thus there is little decoding overhead. \tablename~\ref{compression} shows the address length and the table size for each code employed for the experiments.

We only compress the weight matrix of fully-connected layers in FCDNNs and CNNs. Since fully-connected layers of large-sized CNNs usually consume over 90\% of the weight storage, the weight matrix compression is important not only for FCDNNs but also for CNNs.

% Structured pruning for CONV layers has been studied in~\cite{sajidacm2017}, which is for skipping unnecessary computations easily.

The propagation of a fully-connected layer can be represented with a matrix $W$ as
\begin{equation}
\textbf{y}_{k+1} = \phi_{k+1}(\textbf{W}_{k+1}\textbf{y}_{k} + \textbf{b}_{k+1}),
\end{equation}
where a row corresponds to connections for a single output neuron, and a column corresponds to connections from a single input neuron. We group the weights in the same column of $W$ because the sub-vectors in this direction are less correlated. The performance of row/column sub-vector structured sparse networks is compared in Section~\ref{sec:experiments}.

 %hardware
\begin{figure}[]
%\centering
%\begin{center}
\begin{subfigure}{.2\textwidth}
  \centering
  \includegraphics[width=1.0\linewidth,height=0.8\linewidth]{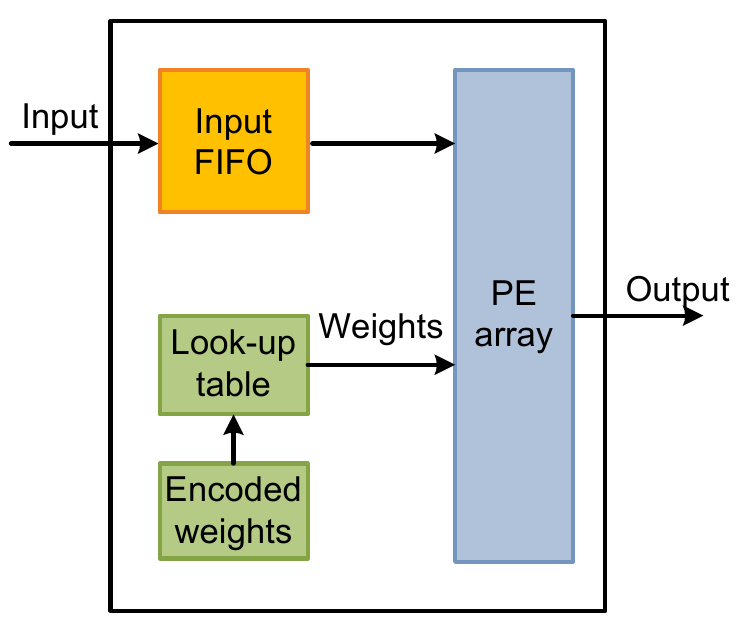}
  \caption{FC layer architecture}
  \label{fig:overall}
\end{subfigure}%
\begin{subfigure}{.3\textwidth}
  \centering
  \includegraphics[width=1.0\linewidth,height=0.6\linewidth]{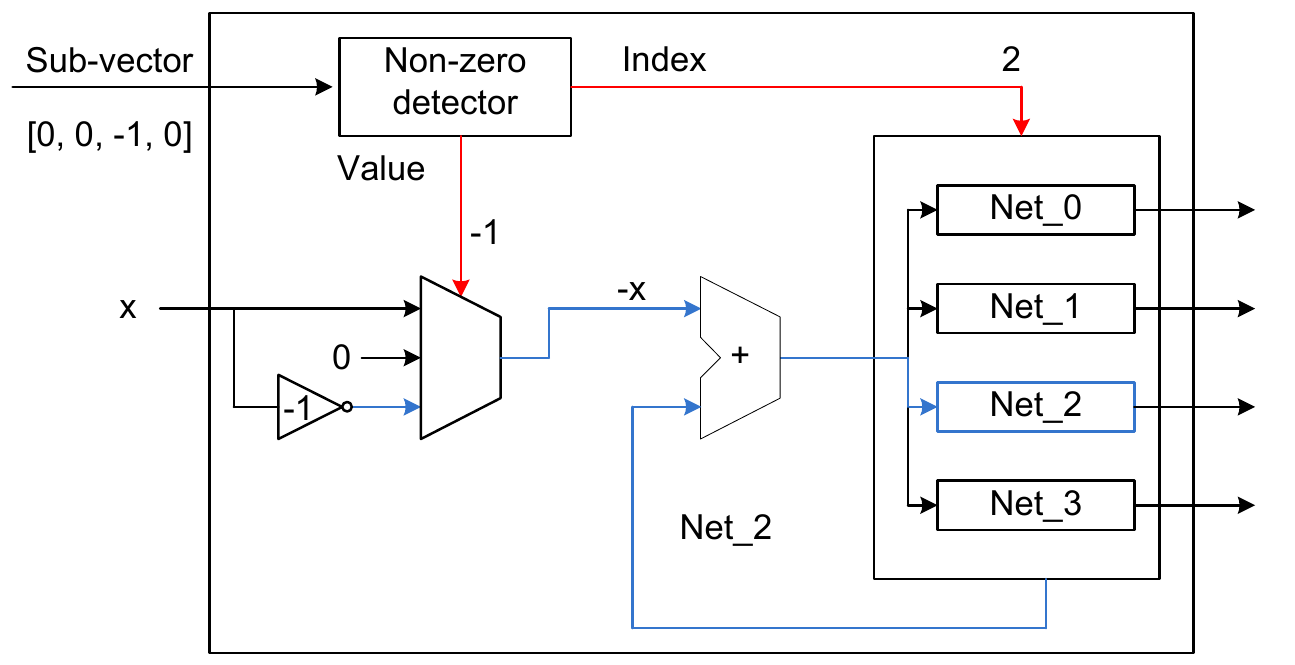}
  \caption{Processing element (PE)}
  \label{fig:pe}
\end{subfigure}

\caption{Architecture of a structured sparse FC layer. (a) Overall architecture. (b) Structure of a processing element when $(N,K)$ is $(4,1)$.}

\label{fig:hardware}
%\end{center}
\end{figure}

\subsection{Hardware design}
\label{subsec:hardware}
 The structured sparse ternary matrix can easily be decoded by employing table look-up operations. The overall architecture for a FC layer can be designed as shown in \figurename~\ref{fig:overall}. A sub-vector index is converted to an uncompressed weight sub-vector simply by indexing the look-up table. Since we drastically reduce the size of encoded weights and the look-up table, as shown in Section~\ref{sec:experiments}, the power consumption by external memory access can be eliminated or reduced with a small weight decoding overhead.

 In addition, unnecessary computation can easily be removed with the column sub-vector structured sparsity. The column sub-vector approach also supports the outer-product based implementation, which is advantageous to parallel processing~\cite{kim2014x1000, park2016fpga}. In the outer product approach, all PEs receive the same input and they conduct outer-product with the consecutive weights. If we choose the column sub-vector based structure, zero weights can be ignored efficiently. \figurename~\ref{fig:pe} shows the structure of PE with an example flow when $(N,K)$ is $(4,1)$. One PE has only $K$ arithmetic units and processes $N$ outputs. For example, when the weight vector is $[0, 0, -1, 0]^{T}$, the non-zero detector block finds the non-zero value $-1$ and the index of non-zero as $2$. The non-zero value detection can easily be done because the length of a sub-vector is small, which is from 4 to 16 in our experiments. Since the weights are ternary quantized, the value just decides whether the input $x$ is added, subtracted, or ignored. The index decides which register is activated. The selected register accumulates the input while others with 0 valued weights keep the current values.

\begin{figure}[]
\begin{center}
	\begin{boxedminipage}{0.49\textwidth}\bfseries
		\raggedright - Masking weights:
		\center{$\boldsymbol{W}_{l} = \boldsymbol{W}_{l,trained}\odot \boldsymbol{M}_{l}$\\}
		\raggedright - Quantization step size determining: \\
		\vspace{5pt}
		\setlength\parindent{50pt}$\Delta_{l} = Qstep(\boldsymbol{W}_{l})$\\
		\vspace{2pt}
		\setlength\parindent{64pt}$ = \operatornamewithlimits{argmin}\limits_{{\Delta}}\operatorname{}\sum\limits_{w_{ij}\in \boldsymbol{W}_{l}}\bigl(Q(w_{ij},\Delta_{l})-w_{ij}\bigr)^2$\\
		\raggedright - Quantized weights:\\
		\vspace{5pt}
		\setlength\parindent{29pt}{$w_{ij}^{(q)} = Q(w_{ij},\Delta)$\\
		\vspace{2pt}
		\setlength\parindent{50pt}$ = sgn(w_{ij}) \cdot \Delta \cdot \min\biggl(\left \lfloor{\dfrac{\lvert w_{ij} \rvert} \Delta +0.5}\right \rfloor, \dfrac{P-1}{2}\biggr)$\\}
		\raggedright - Loss calculation:
		\center{$\boldsymbol{net}_{l} = \boldsymbol{W}_{l}^{(q)}\boldsymbol{y}_{l} + \boldsymbol{b}_{l}$\\}
		\center{$\boldsymbol{y}_{l+1} = \phi_{l}(\boldsymbol{net}_{l})$\\}
		\center{$E = -\sum\limits_{(t,y_{L})}t\log y_{L}$\\}
		\raggedright - Weights update:
		\center{$w_{ij, new} = w_{ij} - \alpha  \dfrac {\partial E}{\partial w_{ij} }\cdot m_{ij}$ \\}
		\center{$w_{ij, new}^{(q)} = Q(w_{ij, new},\Delta_{new})$ \\}
	\end{boxedminipage}
\end{center}

\caption{The retraining algorithm of structured sparse fixed-point network is summarized. $\Delta$ is the quantization step size, $P$ is the number of quantization points, $l$ denotes the layer, which is from 1 to $L$, $\boldsymbol{y}_{l}$ is the output vector of layer $l$, $\phi_{l}()$ is the activation function, $E$ is the loss, $\boldsymbol{t}$ is one-hot encoded label vector, and $\alpha$ is the learning rate. Floating-point weights $\boldsymbol{W}$ and fixed-point weights $\boldsymbol{W}^{(q)}$ are kept pruned by masking matrix $\boldsymbol{M}$.}
\label{fig:algorithm}
\end{figure}

 %algorithm
%\begin{figure}[]
%\centering
%\includegraphics[scale=0.55]{algorithm}
%\caption{The retraining algorithm of structured sparse fixed-point networks is summarized. $\Delta$ is the quantization step size, $n$ is the precision of weights in bits, $net_{i}$ is the summed input value of neuron $i$, $A_{i}$ is the set of neurons anterior to neuron $i$, $P_{j}$ is the set of neurons posterior to neuron $j$, $\phi_{i}( )$ is the activation function, and $\alpha$ is the learning rate. Floating-point weights $W$ and fixed-point weights $W^{(q)}$ are kept pruned by masking matrix $M$.}
%\label{fig:algorithm}
%\end{figure}

\section{Training of structured sparse ternary networks}
\label{sec:retraining}
Structured sparse ternary coding applies a constraint on the maximum number of non-zero weights in addition to ternary quantization. These constraints can incur performance loss. This section explains the structured sparse training method.

\subsection{Structured pruning and quantization}
\label{subsec:algorithm}

%Fixed-point quantization and the network pruning  need a retraining to recover the performance\cite{hwang2014fixed,han2015learning}.

 The network is trained in floating-point first. Then, we simultaneously conduct the structured pruning and quantization, and then retrain the network. Each sub-vector of the floating-point weight matrix is pruned in order of magnitude so that every sub-vector only has $K$ non-zero elements. The masking matrix $M$ is a Boolean matrix showing the pruned locations. In other words, if $w_{ij}$ is pruned, then $m_{ij}$ becomes 0, otherwise 1. The quantization step size $\Delta$ is calculated using the pruned weight matrix instead of the original one. Forward propagation procedure is the same with the previous retraining based quantization algorithm~\cite{hwang2014fixed}.

 We modify the backpropagation algorithm to maintain the structured sparsity during retraining. The gradients for pruned connections are removed by using the masking matrix. This algorithm keeps pruned weights unchanged, while updating the quantized weights. The overall algorithm is illustrated in~\figurename~\ref{fig:structure} and also summarized in~\figurename~\ref{fig:algorithm}.

%histograms
%\begin{figure}[]
%%\centering
%\begin{center}
%\begin{subfigure}{.25\textwidth}
%  \centering
%  \includegraphics[width=.95\linewidth,height=0.5\linewidth]{base_hist}
%  \caption{Baseline(no normalizers)}
%  \label{subfig:base_hist}
%\end{subfigure}%
%\begin{subfigure}{.25\textwidth}
%  \centering
%  \includegraphics[width=.95\linewidth,height=0.5\linewidth]{bn_hist}
%  \caption{BN}
%  \label{subfig:bn_hist}
%\end{subfigure}
%\begin{subfigure}{.25\textwidth}
%  \centering
%  \vskip 0.2in
%  \includegraphics[width=.95\linewidth,height=0.5\linewidth]{wn_hist}
%  \caption{WN before $L_{2}$-normed}
%  \label{subfig:wn_hist}
%\end{subfigure}%
%\begin{subfigure}{.25\textwidth}
%  \centering
%  \vskip 0.2in
%  \includegraphics[width=.95\linewidth,height=0.5\linewidth]{wn_hist_normed_range_0_4}
%  \caption{WN after $L_{2}$-normed}
%  \label{subfig:wn_norm_hist}
%\end{subfigure}
%\caption{Weight distributions trained with various DNN normalizers. The training is performed with InfiMNIST. (a) is trained without any normalizer, (b) is trained with BN, and (c-d) are trained with WN and represent the trained weights $w$ and $L_{2}$-normed weights $v$, respectively.}
%
%\label{fig:hist}
%\end{center}
%\end{figure}

\subsection{Weight distribution normalization with DNN normalizers}
\label{subsec:normalize}
In order to reduce the performance degradation of the structured sparse ternary coding, we apply the batch normalization and weight normalization techniques.

\subsubsection{Batch normalization}
\label{subsubsec:bn}
 Batch normalization (BN) is widely used for training of FCDNNs and CNNs\cite{ioffe2015batch}. BN mitigates the gradient descent problem and acts as a regularizer when the training data is large enough. It normalizes the output neurons of DNNs in the same mini-batch during the forward and backward passes. In each training batch, BN renders the outputs of each neuron follow the Gaussian distribution.

 %\figurename~\ref{subfig:bn_hist} shows the weight distribution of the trained network with BN. It is smoother than the weight distribution of the network trained without a normalizer shown in \figurename~\ref{subfig:base_hist}. Unlike threshold-based unstructured pruning, structured pruning may remove weights that have large magnitude. If the weight distribution is smooth, the performance degradation by pruned connections can be compensated with other connections more easily.

\subsubsection{Weight normalization}
\label{subsubsec:wn}

Weight normalization (WN) uses a simple reparameterization of the weights that accelerates the DNN training~\cite{salimans2016weight}. During training, each weight vector $w$ is rescaled to $v$, which has unit $L_{2}$-norm. The loss is calculated with $v$ and gradients are applied to $w$. For the inference, $L_{2}$-normed weights are used instead of the original ones.

% \figurename~\ref{subfig:wn_hist} and~\ref{subfig:wn_norm_hist} show the weight distributions of $w$ and $v$ trained with WN. WN also makes the weight distribution smooth. Further, $L_{2}$-norm decreases overall weight values. Therefore, WN can also be used to reduce the performance degradation. 

\subsection{Gradual pruning and quantization}
\label{subsec:gradual}

If many connections are pruned in a single step, it is difficult to compensate for the loss by retraining. Gradual pruning can alleviate this problem, which repeats pruning and retraining gradually from low sparsity to high sparsity. At each iteration, a small number of connections are pruned to preserve the performance. Then, after retraining, an increased number of weights are forced to zero and retraining is performed again. In our scheme, we combine the gradual pruning and quantization. At the first iteration, the network is retrained according to the proposed algorithm with a low sparse $(N,K)$, which means a large $K$. Floating-point weights $W$ and fixed-point weights $W^{(q)}$ are both retrained at this iteration. At the next iteration, instead of $W^{(q)}$, $W$ is pruned with higher sparsity by decrementing $K$. Further, we determine the step size $\Delta$ using $W$ because the weight values change much during retraining. The effect of adapting the step size to the changed weights is shown in~\cite{shin2017fixed}. After the network is pruned for the target sparsity, the final $W^{(q)}$ is used for the inference.
 
 % We successfully increase the performance of structured sparse fixed-point AlexNet using the gradual pruning and quantization technique.

\section{Experimental results}
\label{sec:experiments}

\subsection{InfiMNIST}
\label{subsec:mnist}
The MNIST is a handwritten digit recognition dataset that consists of 28$\times$28 greyscale images. The InfiMNIST dataset is derived from the MNIST using pseudo-random deformations and translations~\cite{infimnist2017}. The training set is composed of 1M examples among the 8M sample data. The test set is the same with the original MNIST dataset, and 50K examples in the training set are used for validation. We train the networks using ADAM~\cite{KingmaB14} optimizer. The learning rate decreases from 1e-3 to 1.6e-5 with a factor of 0.2 when the validation does not show improvements for 4 consecutive evaluations. The experimental results are the averages of 5 experiments with different random seeds. The network configuration is as shown below.
%\begin{equation}
%Input(784) - Hidden1(1024) - Hidden2(1024) - 10Softmax,
%\end{equation}
\begin{flalign}
&Input-Hidden1(1024)-Hidden2(1024)-10Softmax,
\end{flalign}

The weight matrix of the output layer is quantized, but not pruned because the size is small. Also, biases and normalization parameters are kept in high precision. The performances according to the direction of the sub-vectors and normalizers are shown in \tablename~\ref{infimnist_results}. In all experiments, the sub-vector length $N$ is 16 and the maximum number of non-zero in a sub-vector $K$ is 3. The ternary network is retrained using the algorithm shown in~\cite{hwang2014fixed}. BN and WN show very high accuracy on the floating-point network because they act as regularizers. Also, BN and WN improve the performance of ternary and structured sparse networks. These results show that the normalizers are effective in alleviating the performance degradation due to the structured sparsity constraint. As we discussed in Section~\ref{sec:SSFN}, the row sub-vector structure results in high error rate. With BN and the column sub-vector structure, we obtain x1.7 times of weight storage compression with 0.06\% miss classification rate (MCR) loss when compared to the unconstrained ternary network.

%infimnist
\begin{table}[]
\centering
\caption{Miss classification rate(MCR(\%)) on the test set with InfiMNIST example. `Baseline' means the networks are trained without any DNN normalizer.}
\label{infimnist_results}
{\renewcommand{\arraystretch}{2}

\setlength{\tabcolsep}{15pt}
\begin{tabular}{c*{3}{c}}
\hline\hline
 & \textbf{Baseline} & \textbf{BN} & \textbf{WN} \\ \hline\hline
Float network    & 1.03              & 0.72        & 0.77        \\ \hline
Ternary network  & 1.30              & 0.86        & 1.05        \\ \hline
Col sub-vector (16,3)  & 1.60              & 0.92        & 1.00        \\ \hline
Row sub-vector (16,3)  & 2.62              & 0.93        & 1.03        \\ \hline\hline
\end{tabular}}
\end{table}

% Another notable point is that the performance of the ternary network trained and retrained with WN decreases much. WN calculates the loss using $L_{2}$-normed weight vector. In the retraining procedure, the loss is computed with $L_{2}$-normed fixed-point weights and the gradients are applied to floating-point weights. However, because of the square operation, $L_{2}$-norm of the floating-point weight vector and the fixed-point weight vector have biased error. Therefore, the ternary network with WN shows poor performance compared to the float-point network. Similar results are also observed in CIFAR-10 experiments.

\subsection{CIFAR-10}
\label{subsec:cifar-10}
The CIFAR-10 dataset includes examples from ten classes: airplane, automobile, bird, cat, deer, dog, frog, horse, ship and truck. The training set consists of 50K 32$\times$32$\times$3 RGB samples and the test set contains 10K samples. We use 10\% of the training set as the validation set. For data augmentation, training data are horizontally flipped with a probability of 50\% at every epoch. Also, global contrast normalization(GCN) is applied to all images. The training procedure is the same with the InfiMNIST task. We use VGG-9, which is modified to accommodate CIFAR-10 input data. The network configuration is as shown below.

%\begin{equation}
%\begin{flalign}
%&Input-(2\times128C3)-MP2-(2\times256C3)-MP2&&\\\nonumber
%&-(2\times512C3)-MP2-(2\times1024FC)-10Softmax&&
%\end{flalign}
%\end{equation}

\begin{flalign}
&Input-(2\times128C3)-MP2-(2\times256C3)-MP2&&\\\nonumber
&-(2\times512C3)-MP2-(2\times1024FC)-10Softmax,&&
\end{flalign}

This network demands 54.8MB for weight storage with the floating-point format and 3.45MB when the network is ternary quantized. For the experiments, BN is applied to the CONV layers and different normalizers are applied to the FC layers. The performances of the networks are shown in \tablename~\ref{cifar10_results_1}. $(N,K)$ is $(16,4)$ for the structured sparse networks. Floating-point networks show the lowest error rate when normalizers are not applied. Although ternary quantization decreases the performance a lot, the column sub-vector structured sparse network trained with BN shows quite good performance. Further, our scheme shows better results than the ternary network when BN and WN are applied. This is because proper pruning can prevent over-fitting.

% The difference between the floating-point network and the column sub-vector structured sparse network of baseline is 1.75\%, whereas BN and WN have 0.37\% and 0.25\% performance losses respectively.

%cifar-10_1
\begin{table}[]
\centering
\caption{MCR(\%) of CIFAR-10 classification task.}
\label{cifar10_results_1}
{\renewcommand{\arraystretch}{2}

\setlength{\tabcolsep}{15pt}
\begin{tabular}{c*{3}{c}}
\hline\hline
 & \textbf{Baseline} & \textbf{BN} & \textbf{WN} \\ \hline\hline
Float network    & 8.00              & 8.55        & 9.16        \\ \hline
Ternary network  & \textbf{9.36}              & 8.94        & 9.74        \\ \hline
Col sub-vector (16,4)  & 9.81              & \textbf{8.92}        & \textbf{9.41}        \\ \hline
Row sub-vector (16,4)  & 9.75              & 8.97        & 9.77        \\ \hline\hline
\end{tabular}}
\end{table}

%cifar-10_2
\begin{table}[]
\centering
\caption{MCR and the weight storage comparison with various $(N,K)$ for CIFAR-10. $N$ is sub-vector length, and $K$ is the number of non-zeros in a sub-vector.}
\label{cifar10_results_2}
{\renewcommand{\arraystretch}{1.5}

\setlength{\tabcolsep}{10pt}
\begin{tabular}{ccccc}
\hline\hline
\multirow{2}{*}{\textit{\textbf{(N,K)}}} & \multicolumn{2}{c}{\textbf{MCR(\%)}} & \multirow{2}{*}{\textbf{\begin{tabular}[c]{@{}c@{}}Weight\\ storage(MB)\end{tabular}}} & \multirow{2}{*}{\textbf{\begin{tabular}[c]{@{}c@{}}Compression\\  ratio\end{tabular}}} \\ 
                                         & \textbf{BN}       & \textbf{WN}        &                                                                                               &                                                                                        \\ \hline\hline
Float                            & 8.55              & 9.16               & 54.804                                                                                        & X1                                                                                     \\ \hline
Ternary                          & 8.94              & \textbf{9.74}   & 3.453                                                                                         & X15.87                                                                                 \\ \hline
(16,4)                                   & 8.92              & 9.87               & 2.433                                                                                         & X22.52                                                                                 \\ \hline
(8,2)                                    & \textbf{8.88}     & 9.78               & 2.301                                                                                         & X23.81                                                                                 \\ \hline
(4,1)                                    & 9.02              & 10.01              & 2.301                                                                                         & X23.81                                                                                 \\ \hline
(16,3)                                   & 9.35              & 10.11              & 2.097                                                                                        & X26.14                                                                                 \\ \hline
(16,2)                                   & 8.93              & 10.36              & 1.874                                                                                       & X29.25                                                                                 \\ \hline
(8,1)                                    & 8.91              & 10.32              & \textbf{1.869}                                                                                & \textbf{X29.32}                                                                        \\ \hline\hline
  
\end{tabular}}
\end{table}

\tablename~\ref{cifar10_results_2} shows the performance and the weight storage compression ratio of networks with various choices of $(N,K)$. The column sub-vector structure is applied to all experiments. When the networks have the same sparsity, the constraint is stronger when $N$ is small, which can decrease the performance. However, if $N$ is large, the weight storage occupied by the table increases. By applying BN and the $(8,1)$ structured sparse network, we achieve x29.32 weight storage compression with MCR loss of 0.36\% compared to the float-point network.

\subsection{ImageNet}
\label{subsec:imagenet}
The ImageNet is a 1000 objects classification problem. We train AlexNet with ImageNet ILSVRC-2012 dataset, which has 1.2M training data and 50K validation data. BN is applied to all CONV layers instead of skipping response normalization. Training is performed with the Matconvnet framework~\cite{vedaldi2015matconvnet}. 

AlexNet has 61M parameters, of which 54.5M parameters are devoted to the FC layers. Therefore, it is important to compress FC layers to reduce the weight storage. The result of retraining to have the structured sparsity for various kinds of $(N,K)$ is shown in \tablename~\ref{imagenet_results_1}. Top-1 and top-5 MCR of the floating-point network are 41.73\% and 18.94\%, respectively. Fixed-point networks are quantized with 8-bit precision for CONV layers and 2-bit precision for FC layers. The unstructured fixed-point network shows 42.37\% top-1 MCR and 19.50\% top-5 MCR. For example, when $(N,K)$ is $(16,4)$, the weight storage compression of x22.64 is obtained compared to the floating-point network. In this case, top-1 error increases 0.43\%. Performance degradation increases as the sparsity grows. By allowing a top-1 error increase of up to 2\% over the floating-point network, x31.68 compression ratio is achievable.

%imagenet
\begin{table}[]
\centering
\caption{Top-1 error rate, top-5 error rate(\%) and the weight storage comparison with various $(N,K)$ for ImageNet classification problem.}
\label{imagenet_results_1}
{\renewcommand{\arraystretch}{1.5}

\setlength{\tabcolsep}{10pt}
\begin{tabular}{ccccc}
\hline\hline
\textit{\textbf{(N,K)}} & \textbf{Top-1} & \textbf{Top-5} &  \textbf{\begin{tabular}[c]{@{}c@{}}Weight\\ storage(MB)\end{tabular}} & \textbf{\begin{tabular}[c]{@{}c@{}}Compression \\ ratio\end{tabular}} \\ \hline\hline
Float           & 41.73                    & 18.94              & 232.61                                                                       & X1                                                                    \\ \hline
Fixed           & 42.37                    & \textbf{19.50}                                                                                      & 16.28                                                                        & X14.29                                                                \\ \hline
(16,4)                  & \textbf{42.17}           & 19.89                                                                                     & 9.92                                                                        & X23.45                                                                \\ \hline
(8,2)                   & 42.58                    & 19.79                                                                                     & 9.78                                                                         & X23.78                                                                \\ \hline
(4,1)                   & 43.04                    & 20.00                                                                                  & 9.78                                                                         & X23.78                                                                \\ \hline
(16,3)                  & 42.94                    & 20.08                                                                                   & 8.78                                                                         & X26.49                                                                \\ \hline
(16,2)                  & 43.51                    & 20.52                                                                               & 7.35                                                                         & X31.66                                                                \\ \hline
(8,1)                   & 43.64                    & 20.66                                                                                & \textbf{7.34}                                                                & \textbf{X31.68}                                                       \\ \hline\hline

\end{tabular}}
\end{table}

%gradual
\begin{table}[]
\centering
\caption{Comparison of the error rate(\%) between the gradual scheme and direct pruning. Note that `Direct' means that the network is pruned by target sparsity at once. All results are evaluated after the retraining is conducted.}
\label{imagenet_results_2}
{\renewcommand{\arraystretch}{2}

\setlength{\tabcolsep}{16pt}
\begin{tabular}{ccccc}
\hline\hline
\multirow{2}{*}{\textit{\textbf{(N,K)}}} & \multicolumn{2}{c}{\textbf{Gradual}}               & \multicolumn{2}{c}{\textbf{Direct}}                \\ \cline{2-3} \cline{4-5}
                                         & \textbf{Top-1} & \textbf{Top-5} & \textbf{Top-1} & \textbf{Top-5} \\ \hline\hline
(8,4)                                    & 42.24                    & 19.75                    & 42.24                    & 19.75                    \\ \hline
(8,3)                                    & \textbf{42.05}                    & \textbf{19.68}                    & 42.65                    & 19.93                    \\ \hline
(8,2)                                    & \textbf{42.48}                    & 19.92                    & 42.58                    & \textbf{19.79}                    \\ \hline
(8,1)                                    & \textbf{43.07}                    & \textbf{20.26}                    & 43.64                    & 20.66                    \\ \hline\hline

\end{tabular}}
\end{table}

We further perform the gradual pruning to increase the performance of the structured sparse ternary network. The results are shown in \tablename~\ref{imagenet_results_2}. For every iteration, $K$ is decremented by 1. Gradually pruned network shows 43.07\% top-1 MCR, which is lower than that of the directly pruned network by 0.57\%. Finally, we obtain x31.68 weight storage compression ratio with a top-1 MCR increase of 1.34\% over the floating-point network.

We compare our model with other fixed-point networks on \tablename~\ref{comparison}. The baseline is the floating-point network trained using Matconvnet framework. XNOR-Net achieves a compression ratio of x32 through binary quantization, but the error increases sharply. Ternary weight network (TWN)~\cite{li2016ternary} and TWN V2~\cite{zhu2017trained} quantize weights to ternary values. TWN V2 shows 42.5\% Top-1 error rate, which is the lowest among the fixed-point networks. Deep Compression uses the relative indexed CSC/CSR format and Huffman coding to compress the sparse fixed-point network and reduces the weight storage by x35 while maintaining the performance. However, TWN V2 and Deep Compression have disadvantages of hardware implementation. Since each quantization point has a different step size, high-precision arithmetic operations are needed. On the other hand, XNOR-Net, TWN, and ours can substitute the multiply-accumulate operations to simple logical or accumulate operations~\cite{rastegari2016xnor, park2016fpga, kim2014x1000}. Further, Deep Compression needs additional decoding units~\cite{han2016eie}. Considering the complexity and overhead, our coding scheme is very advantageous to hardware implementations.

%comparison
\begin{table}[]
\centering
\caption{Comparison of networks in the error rate(\%) and the weight storage compression ratio. Results are on ImageNet data with AlexNet.}
\label{comparison}
{\renewcommand{\arraystretch}{1.5}

\setlength{\tabcolsep}{5pt}
\begin{tabular}{lcccc}
\hline\hline
 & \begin{tabular}[c]{@{}c@{}}\textbf{Top-1}\end{tabular} & \begin{tabular}[c]{@{}c@{}}\textbf{Top-5}\end{tabular} & \begin{tabular}[c]{@{}c@{}}\textbf{Weight}\\ \textbf{storage(MB)}\end{tabular} & \textbf{Ratio} \\ \hline\hline
Float(baseline)            & 41.7  & 18.9  & 232.6                                                         & X1    \\ \hline
XNOR-Net         & 55.8  & 30.8  & 7.3                                                           & X32   \\ \hline
TWN              & 45.5  & 23.2  & 17.0                                                          & X14   \\ \hline
TWN\_V2          & \textbf{42.5}  & 20.3  & 17.0                                                          & X14   \\ \hline
Deep Compression & 42.8  & \textbf{19.7}  & \textbf{6.9}                                                           & \textbf{X35}   \\ \hline
\textbf{Ours}             & 43.1  & 20.3  & 7.3                                                           & X32   \\ \hline\hline
\end{tabular}}
\end{table}

\section{Concluding remarks}
\label{sec:conclusion}
We presented a structured sparse ternary coding scheme for low-energy hardware implementation of deep neural networks. The proposed method compresses a matrix for a fully-connected layer by decomposing it into sub-vectors and allowing only a limited number of non-zero values in each sub-vector. The decoding is conducted simply by consulting a look-up table.  Batch normalization and gradual pruning techniques are employed to mitigate the performance degradation due to structured sparsity. We can reduce the weight storage for VGG-9 and AlexNet to 1.87MB (x29 compression) and 7.34MB (x32 compression) with only small accuracy loss. This research is useful for the implementation of large-sized DNNs on resource-limited hardware.

% conference papers do not normally have an appendix

% use section* for acknowledgment
\section*{Acknowledgment}
\label{ack}
This work was supported in part by the Brain Korea 21 Plus Project and the National Research Foundation of Korea (NRF) grant funded by the Korea government (MSIP) (No. 2015R1A2A1A10056051).

\bibliographystyle{IEEEbib}
\bibliography{refs}

% that's all folks
\end{document}